\documentclass[letterpaper, 10 pt, journal, twoside]{IEEEtran}
\usepackage{amsmath}       
\usepackage{amssymb}       
\usepackage{amsfonts}      
\usepackage{mathtools}     
\usepackage{bm}            
\usepackage{bbm}           
\usepackage{xfrac}
\usepackage{xspace}        
\usepackage{soul}
\usepackage[normalem]{ulem} 

\usepackage{graphicx}      
\usepackage{wrapfig}       
\usepackage{float}         
\usepackage{stfloats}      
\usepackage{capt-of}       
\usepackage{subfigure}
\usepackage[font=footnotesize,labelfont=bf]{caption}      

\usepackage{booktabs}      
\usepackage{multirow}      
\usepackage[flushleft]{threeparttable} 
\usepackage{arydshln}      
\usepackage{listings}
\usepackage[dvipsnames,table,xcdraw]{xcolor} 

\definecolor{mydarkblue}{rgb}{0,0.08,0.45}
\definecolor{mydarkgreen}{RGB}{0, 139, 69}
\definecolor{mygreen2}{RGB}{0, 205, 0}
\definecolor{mybrown}{RGB}{139, 69, 19}
\definecolor{boxblue}{RGB}{79,173,234}
\definecolor{tablepeach}{RGB}{255, 240, 235}
\definecolor{tablepurple}{RGB}{248,235,252}
\definecolor{tableblue}{RGB}{235,241,255}

\usepackage[bookmarks=true]{hyperref}         
\usepackage[capitalise, nameinlink]{cleveref} 
\usepackage{xurl}                       
\definecolor{citecolor}{HTML}{3A33FF} 

\hypersetup{
    colorlinks=true,
    linkcolor=citecolor,
    urlcolor=citecolor,
    citecolor=citecolor,
}
\usepackage[linesnumbered,ruled,vlined]{algorithm2e} 

\usepackage{pifont}  
\usepackage{dsfont}  
\usepackage{lipsum}  
\usepackage{siunitx} 
\DeclareSIUnit\inch{in}
\usepackage{multicol} 
\let\labelindent\relax
\usepackage{enumitem}

\newcommand{\ci}[1]{\scriptsize{\textcolor{gray}{~($\pm #1$)}}}

\newcommand{\cilmit}[1]{\cilmitparse#1\cilmitend}
\def\cilmitparse#1,#2,#3\cilmitend{\scriptsize{\textcolor{gray}{~$\left(\pm #1,\ \substack{\scriptscriptstyle #2\\\scriptscriptstyle #3}\right)$}}}

\begin{document}
\bstctlcite{IEEEtranBSTCTL:etal} 
\title{Learning End-to-End Control for Omnidirectional Aerial Motion on Overactuated Tilt-rotor Quadrotors}
%
%
%

\author{Wentao Zhang$^{1}$, Zhaoqi Ma$^{1}$, Jinjie Li$^{1}$, Huayi Wang$^{2}$, Haokun Liu$^{1}$, Junichiro Sugihara$^{1}$, Chen Chen$^{1}$, Yicheng Chen$^{1}$, Cuniato Eugenio$^{3}$, Pantic Michael$^{3}$ and Zhao, Moju$^{1}$%
\thanks{Manuscript received May, 30, 2026; Revised August, 24, 2026; Accepted September, 12, 2026. This work was supported by JST ASPIRE (JPMJAP2521) and in part by CSC. (Corresponding author: Moju Zhao.)} 
\thanks{$^{1}$Wentao Zhang, Zhaoqi Ma, Jinjie Li, Haokun Liu, Junichiro Sugihara, Chen Chen, Yicheng Chen and Moju Zhao are with the DRAGON Lab, Department of Mechanical Engineering, The University of Tokyo, Tokyo, 113-8656, Japan
        {\tt\footnotesize \{wentao-zhang, zhaoqi-ma, jinjie-li, haokun-liu, j-sugihara, chen-chen, yicheng-chen, chou\}@dragon.t.u-tokyo.ac.jp}}
\thanks{$^{2}$Huayi Wang is with the School of Computing and Data Science, The University of Hong Kong, 999077, HKSAR
        {\tt\footnotesize huayiwang@connect.hku.hk}}%
\thanks{$^{3}$Cuniato Eugenio and Pantic Michael are with the Autonomous Systems Lab(ASL), ETH Zurich, 8092 Zurich, Switzerland
        {\tt\footnotesize \{ecuniato, mpantic\}@ethz.ch}}%
\thanks{Digital Object Identifier (DOI): see top of this page.}
}

\IEEEaftertitletext{%
\begin{center}
    \captionsetup{type=figure}
    \vspace{-0.6cm}
    \includegraphics[width=0.9\textwidth]{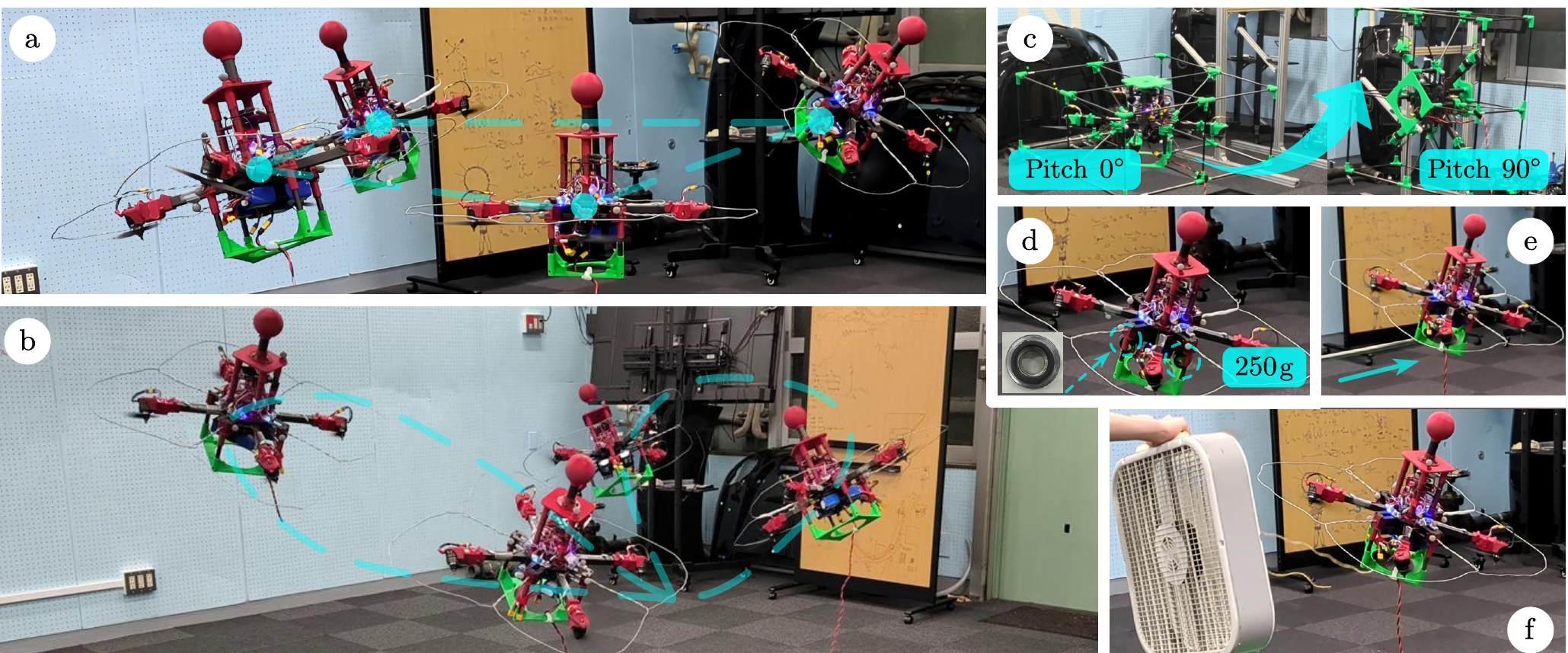}
    \caption{Proposed end-to-end RL framework enables omnidirectional real-world flight across multiple scenarios. (a) Waypoint hovering, (b) Trajectory tracking, (c) Vertical tilt angle operation, (d) Payload variation (\SI{0.25}{\kilogram} $\times$ 2), (e) External force recovery, and (f) Wind disturbance rejection (\SI{2.5}{\metre\per\second}).}\label{fig:highlight}
    \vspace{-0.2cm}
\end{center}
}

\maketitle
\begin{abstract}
While reinforcement learning (RL) has been successfully applied to conventional quadrotors for agile and robust flight, whether actuator-level RL can be reliably deployed on tilt-rotor aerial robots remains an open question, as the hybrid actuation coupling brushless rotors with rotational joints introduces a substantially harder sim-to-real gap. In this work, we propose an end-to-end RL framework for omnidirectional motion control on overactuated tilt-rotor quadrotors, directly mapping target poses to joint and rotor commands. The learning framework combines actuator-level simulation with an asymmetric actor-critic architecture for 6D pose-reaching. For reliable sim-to-real transfer on the hybrid actuation, we integrate system identification with minimal yet physically grounded domain randomization. The trained policy is deployed zero-shot on real hardware and evaluated across waypoint hovering, external disturbances, payload variation and trajectory tracking, together with simulated traversal of allocation-singular configurations. The policy is compared with a state-of-the-art NMPC baseline: NMPC attains lower steady-state position error, whereas the RL policy offers a more uniform orientation error across evaluations, transitions between poses faster, and requires less onboard computation.
\end{abstract}


%
\IEEEpeerreviewmaketitle
\section{Introduction}\label{sec:introduction}
Aerial robots have demonstrated remarkable capabilities across a wide range of applications. In particular, quadrotors have been extensively deployed in diverse real-world missions, including inspection, delivery, and search and rescue~\cite{floreano2015science}. Despite their success, conventional quadrotors are inherently underactuated and cannot independently regulate position and orientation, which limits their applicability to advanced aerial interaction. To address this limitation, omnidirectional aerial robots have recently attracted growing attention for their ability to perform full six-degree-of-freedom (6D) motion~\cite{li2024fixed,gupta2025umi,allenspach2020design}. Among the various designs, tilt-rotor platforms stand out by actively reorienting individual thrust vectors, attaining high actuation efficiency without internal force cancellation~\cite{allenspach2020design,zhao2023design,zhao2023versatile}. However, this expanded capability comes at the cost of substantially increased system complexity. The strong coupling between joint and rotor dynamics, combined with the highly flexible omnidirectional motion capability (Fig.~\ref{fig:highlight}), poses fundamental challenges for motion control.

\begin{figure}[htbp]
    \vspace{-0.3cm}
    \centering
    \includegraphics[width = 8.0 cm]{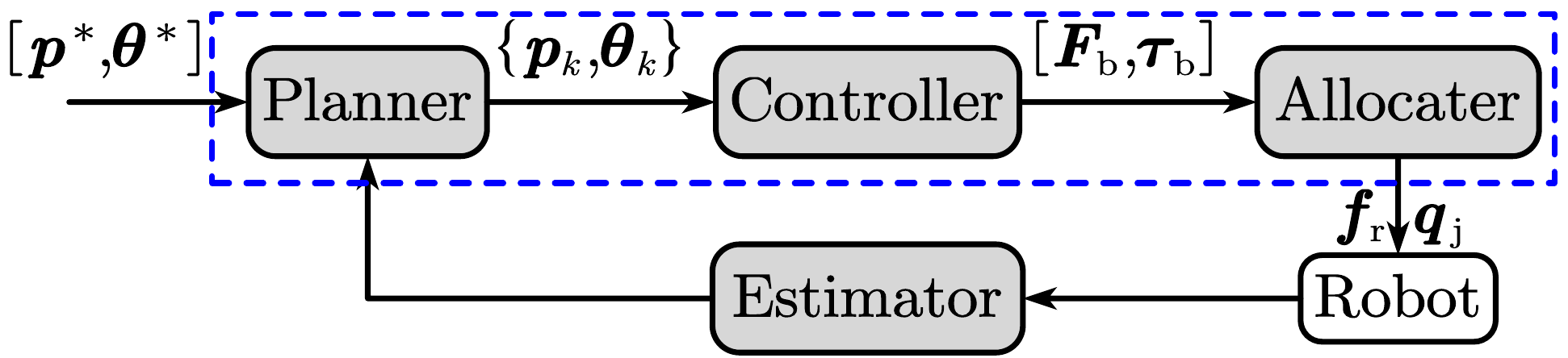}
    \vspace{-0.1cm}
    \caption{Model-based control framework for tiltable aerial robots.}
    \label{fig:modelbased}
    \vspace{-0.3cm}
\end{figure}

Model-based control has long been the mainstream approach for tilt-rotor aerial robots, with geometric control~\cite{allenspach2020design,lee2025autonomous,sugihara2024beatle} and Model Predictive Control (MPC)~\cite{li2024servo,li2025six} as the most representative formulations. These methods typically adopt a sequential structure, as illustrated in Fig.~\ref{fig:modelbased}. For an aerial robot expressed in the body frame with mass $m$, inertia $\boldsymbol{J}_{\mathrm{b}}$, linear velocity $\boldsymbol{v}_{\mathrm{b}}$, and angular velocity $\boldsymbol{\omega}_{\mathrm{b}}$, the rigid-body dynamics are written as:
\begin{equation}
    \begin{aligned}
    \left[\!\! \begin{array}{c}
        \boldsymbol{F}_{\mathrm{b}}\\
        \boldsymbol{\tau }_{\mathrm{b}}\\
    \end{array}\!\!\right]\!+\!\left[\!\!\begin{array}{c}
        \boldsymbol{F}_{\mathrm{d}}\\
        \boldsymbol{\tau }_{\mathrm{d}}\\
    \end{array}\!\!\right]\!=\!\left[\!\!\begin{array}{c}
        m\boldsymbol{I}_{\mathrm{3}}\,\,\boldsymbol{0}\\
        \boldsymbol{0}\,\,\,\,\,\,  \boldsymbol{J}_{\mathrm{b}}\\
    \end{array}\!\!\right]\!\left[\!\!\begin{array}{c}
        \dot{\boldsymbol{v}}_{\mathrm{b}}\\
        \dot{\boldsymbol{\omega}}_{\mathrm{b}}\\
    \end{array}\!\!\right]\!+\!\left[\!\!\begin{array}{c}
        \boldsymbol{\omega }_{\mathrm{b}}\times m{\boldsymbol{v}}_{\mathrm{b}}\\
        \boldsymbol{\omega }_{\mathrm{b}}\times \boldsymbol{J}_{\mathrm{b}}\boldsymbol{\omega }_{\mathrm{b}}\\
    \end{array}\!\!\right]\label{eq:model}
        \end{aligned}
        \vspace{-0.1cm}
\end{equation}
where $[\boldsymbol{F}_{\mathrm{b}}\,\boldsymbol{\tau}_{\mathrm{b}}]^{\top}$ denotes the body wrench produced by the actuators and $[\boldsymbol{F}_{\mathrm{d}}\,\boldsymbol{\tau}_{\mathrm{d}}]^{\top}$ represents the external disturbance wrench. For tilt-rotor platforms, the body wrench is jointly produced by joint positions $\boldsymbol{q}_{\mathrm{j}}$ and rotor thrusts $\boldsymbol{f}_{\mathrm{r}}$, coupled through the allocation matrix $\boldsymbol{A}_{\alpha}$:
\vspace{-0.1cm}
\begin{equation}
\left[ \begin{matrix}
	\boldsymbol{F}_{\mathrm{b}}\!&		\!\boldsymbol{\tau }_{\mathrm{b}}\\
\end{matrix} \right] ^{\top}=\boldsymbol{A}_{\alpha}\left( \boldsymbol{q}_{\mathrm{j}} \right) \left[ \begin{matrix}
	f_{\mathrm{r},1}&		f_{\mathrm{r},2}&		\cdots\\
\end{matrix} \right] ^{\top}\label{eq:allocation}
\vspace{-0.1cm}
\end{equation}
Actuator commands are then recovered by inverting this allocation mapping. Under nominal conditions, the hierarchical framework achieves accurate pose regulation and offers strong layer-wise interpretability.

Beyond accuracy and interpretability, this decomposition keeps each subproblem solvable at the control rate, enabling onboard deployment. However, several characteristics of model-based methods motivate complementary research directions: (i) the model must stay tractable within the online control period, so it is often reduced under simplifying assumptions; unmodeled dynamics and external disturbances $[\boldsymbol{F}_{\mathrm{d}}\,\boldsymbol{\tau}_{\mathrm{d}}]$ are then handled by dedicated observers and compensated through the allocator~\cite{lee2025autonomous}, whereas learning-based formulations can propagate the dynamic model offline, through large-scale training-time exposure; (ii) the allocation matrix $\boldsymbol{A}_{\alpha}$ becomes ill-conditioned at certain joint configurations, which prior work has addressed through explicit regularization or actuator-dynamics integration~\cite{li2024servo,cuniato2026allocation}. These observations motivate us to investigate whether a policy trained offline can directly operate actuators while subsuming planning, control, and allocation within a single policy.

Learning-based approaches have emerged as a promising alternative to model-based control and have showcased impressive performance across various aerial robotics tasks.~\cite{hwangbo2017control} pioneered actuator-level reinforcement learning (RL) for quadrotor waypoint tracking.~\cite{o2022neural} demonstrated robust disturbance rejection under strong winds via deep learning combined with meta-learning.~\cite{song2023reaching} and~\cite{kaufmann2023champion} achieved aggressive quadrotor flight comparable to human champions in drone racing.~\cite{eschmann2026raptor} trained a single policy that generalizes across quadrotor platforms with diverse physical parameters and rotor configurations. While these studies highlight the potential of learning-based control, they are largely confined to fixed-rotor platforms. To the best of our knowledge, data-driven control for tilt-rotor robots remains significantly underexplored, with only a few works integrating learned components into MPC~\cite{brunner2022mpc}, guiding policy training with a model-based action prior~\cite{cuniato2023learning}, or fine-tuning actuator-level policies with real-world data for sim-to-real transfer~\cite{lee2021low}. These preliminary efforts suggest that learning-based control is promising for tilt-rotor platforms, but also expose the substantial challenges in learning formulation and sim-to-real transfer on such complex systems.

\begin{figure}[htbp]
    \vspace{-0.3cm}
    \centering
    \includegraphics[width = 8.0 cm]{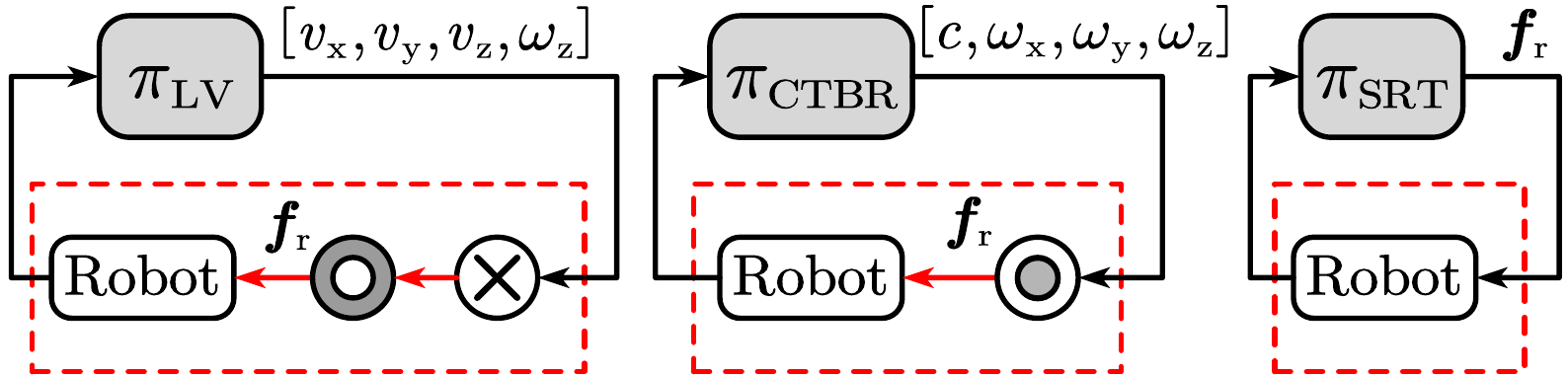}
    \caption{Data-driven control frameworks for quadrotors. LV: rely on a control stack to process velocity commands. CTBR: require a feedback regulator to process rotor thrusts. SRT: directly command rotor thrusts.}
    \label{fig:learningframework}
    \vspace{-0.3cm}
\end{figure}

Adopting RL on a new robotic platform fundamentally requires answering two questions: (1) \textbf{what} learning formulation fits the platform, and (2) \textbf{how} can the trained policy be reliably transferred from simulation to the real system? Prior work on fixed-rotor quadrotors has explored three action-space abstractions that occupy distinct points along the trade-off between sim-to-real transferability and the breadth of system dynamics they can exploit~\cite{kaufmann2022benchmark}, as illustrated in Fig.~\ref{fig:learningframework}. The Linear Velocity (LV) abstraction is the most high-level: it transfers easily with velocity controller but underexploits the platform's full dynamic capabilities. The Collective Thrust and Bodyrates (CTBR) abstraction strikes a balance between transferability and dynamic expressiveness, and is widely adopted on fixed-rotor systems~\cite{kaufmann2023champion,song2023reaching}; however, applying CTBR to tilt-rotor platforms still requires a low-level allocator beneath the policy, leaving the singularity-prone allocation problem unresolved. The Single-Rotor Thrusts (SRT) abstraction operates directly at the actuator level and can holistically capture the platform's dynamics~\cite{hwangbo2017control}, but is less commonly used because of the stronger sim-to-real demands it imposes.

Given the highly flexible motion capability and strong joint-rotor coupling of tilt-rotor platforms, we adopt an actuator-level abstraction analogous to SRT and propose an end-to-end RL framework for 6D pose-reaching control on overactuated tilt-rotor quadrotors. The policy directly maps the target pose and current state to joint position and rotor thrust commands with a single policy. However, this approach introduces a more severe sim-to-real gap, further exacerbated by the hybrid actuation system that couples brushless rotors with servo-driven joints. To bridge this gap, we integrate actuator system identification with minimal yet physically grounded domain randomization into a sim-to-real pipeline, validated through extensive real-world experiments. The main contributions are summarized as follows:

\begin{itemize}
    \item An end-to-end reinforcement learning framework is proposed, which produces joint and rotor commands from a single policy on overactuated tilt-rotor quadrotors, enabling actuator-level 6D pose-reaching control.
    \item A sim-to-real pipeline tailored to the hybrid actuation is introduced, integrating actuator system identification, control-architecture latency alignment, and minimal yet physically grounded domain randomization.
    \item Zero-shot deployment across five real-world flight scenarios and a simulated traversal of allocation-singular configurations, together with a matched comparison against a servo-integrated NMPC baseline on the same platform and software stack, in which the policy holds a more uniform orientation error at lower online computation.
\end{itemize}

\begin{figure*}[htbp]
    \vspace{0.1cm}
    \centering
    \includegraphics[width=0.95\textwidth]{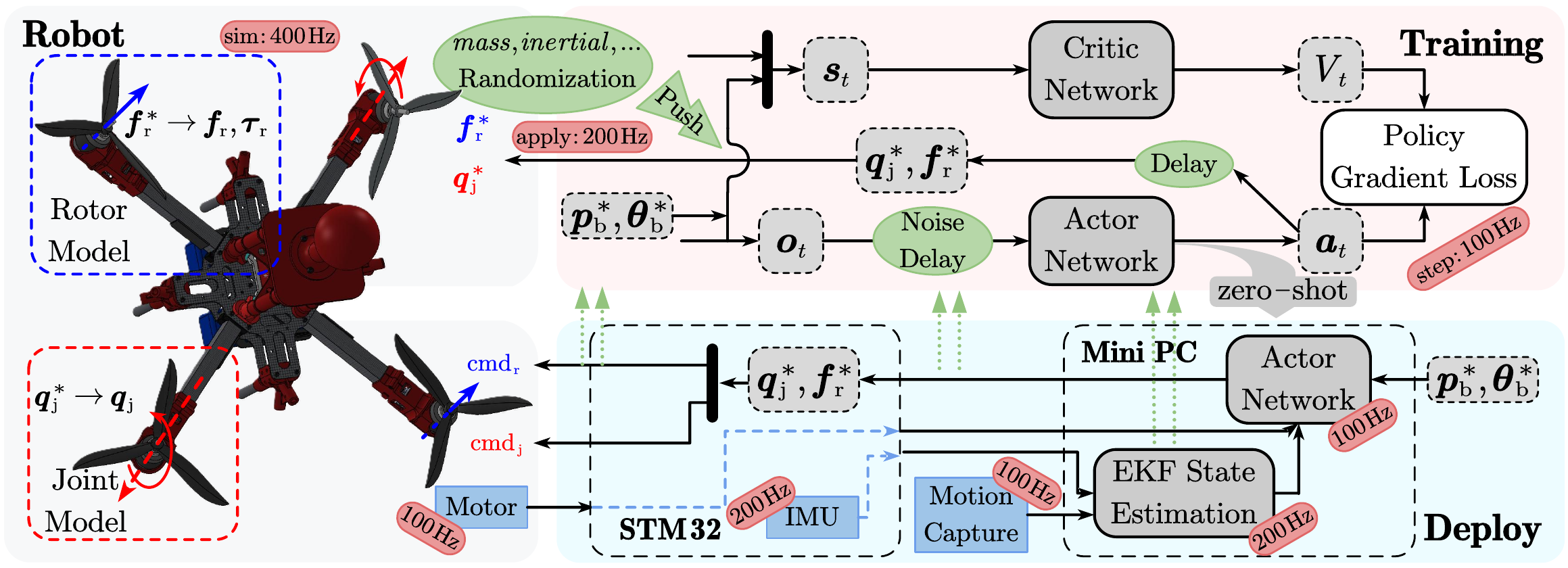}
    \caption{Overall framework for simulation, RL training, and deployment. \textbf{Robot}: Rotor modules (blue, target thrust ${\boldsymbol{f}}_{\mathrm{r}}^{*} \rightarrow$ real thrust $\boldsymbol{f}_{\mathrm{r}}$ and torque $\boldsymbol{\tau}_{\mathrm{r}}$), joint modules (red, joint position commands $\boldsymbol{q}_{\mathrm{j}}^* \rightarrow$ position feedback $\boldsymbol{q}_{\mathrm{j}}$), and system latency (green arrows) are modeled to construct physically consistent simulation environments. \textbf{Training}: The policy is trained under an asymmetric actor-critic framework and dynamics randomization (green) is integrated according to models and deployment scheme. \textbf{Deployment}: Trained policy is deployed on real robot platform with a consistent system architecture. The robot's state is estimated by fusing data from a motion capture system (MoCap) and an onboard IMU using an Extended Kalman filter (EKF).}\label{fig:framework}
    \vspace{-0.5cm}
\end{figure*}
\section{Method}\label{sec:methods}
The proposed RL framework for pose-reaching control is designed in accordance with the real robot platform and deployment scheme, as shown in Fig.~\ref{fig:framework}.
\subsection{Preliminaries}
\subsubsection{Platform Introduction}
The tilt-rotor quadrotor platform consists of four rotor modules, each mounted at the tip of one arm. Each module uses a three-bladed \SI{9}{\inch} propeller driven by a T-Motor AT2814 KV900 brushless DC motor and is connected to the arm through a single-degree-of-freedom rotational joint actuated by a DYNAMIXEL XC330 T181-T servomotor. This design enables full 6-DOF wrench generation; the detailed physical parameters are listed in Appendix~\ref{apd:physical}.

\subsubsection{Task Definition}
The control task is defined as reaching a target pose $\left[\boldsymbol{p}_{\mathrm{w}}^{*}, \boldsymbol{\theta}_{\mathrm{b}}^{*}\right]$ in $SE(3)$ and maintaining stability thereafter, by directly commanding the joint positions $\boldsymbol{q}_{\mathrm{j}}$ and rotor thrusts $\boldsymbol{f}_{\mathrm{r}}$. In contrast to velocity-tracking or body-wrench regulation, this pose-level formulation grants the policy full authority over the transition trajectory, prompting it to discover efficient motion strategies through trial-and-error interactions. 

\subsubsection{Problem Formulation}
The proposed RL control problem is formulated as an infinite-horizon Partially Observable Markov Decision Process (POMDP), defined by the tuple $(\mathcal{S}, \mathcal{O}, \mathcal{A}, \mathcal{T}, d_{0}, r, \gamma)$~\cite{lauri2022partially}. Here, $\boldsymbol{s}\in\mathcal{S}$ denotes the full system state, $\boldsymbol{o}\in\mathcal{O}$ represents the partial observation, and $\boldsymbol{a}\in\mathcal{A}$ is action. The initial state distribution is given by $d_{0}$, and the state transition dynamics are described by $\mathcal{T}(\boldsymbol{s}_{t+1}|\boldsymbol{s}_{t},\boldsymbol{a}_{t})$. The reward function is denoted by $r(\boldsymbol{s}_{t},\boldsymbol{a}_{t})$ with discount factor $\gamma \in [0,1)$. The objective is to learn a policy $\pi_{\varTheta}(\boldsymbol{a}|\boldsymbol{o})$, parameterized by $\varTheta$, that maximizes the expected cumulative discounted reward $\mathbb{E}\left[\sum_{t=0}^{\infty}\gamma^{t}r(\boldsymbol{s}_{t},\boldsymbol{a}_{t})\right]$.

\subsection{Training Scheme}
\subsubsection{Observation and Action Space}
The policy action at time-step $t$ is defined as
\begin{equation}
    \boldsymbol{a}_t = \bigl[\boldsymbol{a}_{\mathrm{j},t}, \boldsymbol{a}_{\mathrm{r},t}\bigr] \in \mathbb{R}^{8},
\end{equation}
where $\boldsymbol{a}_{\mathrm{j},t} \in \mathbb{R}^4$ and $\boldsymbol{a}_{\mathrm{r},t} \in \mathbb{R}^4$ are the joint and rotor-thrust components, respectively. They are mapped to the target joint positions $\boldsymbol{q}_{\mathrm{j},t}^{*}$ and rotor thrusts $\boldsymbol{f}_{\mathrm{r},t}^{*}$ as
\begin{equation}
    \boldsymbol{q}_{\mathrm{j},t}^{*} = c_{\mathrm{j}}\boldsymbol{a}_{\mathrm{j},t}, \quad 
    \boldsymbol{f}_{\mathrm{r},t}^{*} = c_{\mathrm{r}}\boldsymbol{a}_{\mathrm{r},t} + \mathrm{C}_{\mathrm{f}},
\end{equation}
where $c_{\mathrm{j}}$ and $c_{\mathrm{r}}$ are scale coefficients and $\mathrm{C}_{\mathrm{f}}$ is the nominal hovering thrust that approximately balances gravity. Centering the action around $\mathrm{C}_{\mathrm{f}}$ lets the policy issue both positive and negative thrust adjustments about hover, facilitating early-stage exploration and yielding smoother commands. The joint and rotor models are obtained through system identification, as detailed in Sec.~\ref{sec:rotorModule} and Sec.~\ref{sec:jointModule}.

The policy observation $\boldsymbol{o}_t \in \mathbb{R}^{33}$ at time-step $t$ is
\begin{equation}
   \boldsymbol{o}_t = \bigl[\boldsymbol{v}_{\mathrm{b},t}, \boldsymbol{\omega}_{\mathrm{b},t}, \boldsymbol{p}_{\mathrm{b},t}^{*}, R_{\mathrm{b},t}, R_{\mathrm{b},t}^{*}, \boldsymbol{q}_{\mathrm{j},t}, \boldsymbol{a}_{t-1}\bigr],
\end{equation}
where the subscript $_\mathrm{b}$ denotes quantities in the robot base frame. $\boldsymbol{v}_{\mathrm{b},t}, \boldsymbol{\omega}_{\mathrm{b},t} \in \mathbb{R}^3$ are the base linear and angular velocities. $\boldsymbol{p}_{\mathrm{b},t}^{*} \in \mathbb{R}^3$ is the world-frame target $\boldsymbol{p}_{\mathrm{w}}^{*}$ transformed into the current body frame. The current and target orientations $R_{\mathrm{b},t}, R_{\mathrm{b},t}^{*} \in \mathbb{R}^6$ are encoded by the first two columns of their rotation matrices~\cite{zhou2019continuity}, a continuous representation that improves learning stability over quaternion or Euler-angle parameterizations. $\boldsymbol{q}_{\mathrm{j},t} \in \mathbb{R}^4$ contains the feedback joint positions, and $\boldsymbol{a}_{t-1}$ is the previous action.

Since the deployed policy has access only to partial observations, we adopt an asymmetric actor-critic framework~\cite{pinto2018asymmetric}: the actor is conditioned on $\boldsymbol{o}_t$, while the critic is fed with a richer state $\boldsymbol{s}_t \in \mathbb{R}^{41}$ for better value estimation,
\begin{equation}
   \boldsymbol{s}_t = \bigl[\boldsymbol{v}_{\mathrm{b},t}, \boldsymbol{\omega}_{\mathrm{b},t}, \boldsymbol{p}_{\mathrm{b},t}^{*}, R_{\mathrm{b},t}, R_{\mathrm{b},t}^{*}, \boldsymbol{q}_{\mathrm{j},t}, \boldsymbol{f}_{\mathrm{r},t}, \boldsymbol{\tau}_{\mathrm{r},t}, \boldsymbol{a}_{t-1}\bigr],
\end{equation}
where the privileged variables are the rotor thrusts $\boldsymbol{f}_{\mathrm{r},t}$ and the corresponding reaction torques $\boldsymbol{\tau}_{\mathrm{r},t} \in \mathbb{R}^4$.

\subsubsection{Reward Function}\label{sec:reward}
The control objective is to drive the system toward a desired pose while maintaining smooth and stable motion. We characterize desirable performance as fast convergence to the target pose with reduced oscillations and moderate control effort, in line with criteria commonly adopted in model-based control. Accordingly, the reward function consists of: (1) \textit{pose-reaching} terms that encourage rapid convergence toward the desired pose, and (2) \textit{regularization} terms that promote smooth and stable motion. All terms and their weights are summarized in Table~\ref{tab:reward}. 
\begin{table}[htbp]
	\centering
    \renewcommand{\arraystretch}{1.4}
    \setlength{\tabcolsep}{2.5pt}
	\caption{\textbf{Reward Description}}\label{tab:reward}
    \vspace{-0.01in}
    \resizebox{0.485\textwidth}{!}{
	\begin{tabular}{lll}
	\hline
	\textbf{Reward}& \multicolumn{1}{c}{\textbf{Formulation}} & \textbf{Weight} \\  
    \midrule[0.7pt]
    \rowcolor[gray]{0.9} \multicolumn{3}{l}{\textbf{\textit{Pose Reaching}}} \\
    \midrule[0.7pt]
	Towards Position & $\tanh \!\left( \frac{\parallel \boldsymbol{p}_e\parallel}{0.6} \right) $  & $-1.0$\\
    Towards Pose & $\left( 1-\tanh \!\left( \frac{\parallel \boldsymbol{\theta }_e\parallel}{2} \right) \right) \cdot \left( 1-\tanh \!\left( \frac{\parallel \boldsymbol{p}_e\parallel}{2} \right) \right)$ & $2.0$\\
    Reach Position &$\left[ \hspace{-5pt}\begin{array}{c}
	\mathbb{I} \left[ \parallel \boldsymbol{p}_e\parallel <p_{\mathrm{th}} \right]\\
	\land \mathbb{I} \left[ \parallel \boldsymbol{v}_{\mathrm{b}}\parallel <v_{\mathrm{th}} \right]\\
\end{array} \hspace{-5pt}\right] \cdot \left[ e^{-\frac{\parallel \boldsymbol{p}_e\parallel}{p_{\mathrm{th}}}}\hspace{-2pt}+e^{-\frac{\parallel \boldsymbol{v}_{\mathrm{b}}\parallel}{v_{\mathrm{th}}}} \right]$ & $1.0$\\
	Reach Pose &\hspace{-10pt} $\begin{array}{c}
	\left[\hspace{-5pt}\begin{array}{c}
	\mathbb{I} \left[ \parallel \boldsymbol{p}_e\parallel <0.02 \right]\land\mathbb{I} \left[ \parallel \boldsymbol{\theta }_e\parallel <0.05 \right]\\
\land\mathbb{I} \left[ \parallel \boldsymbol{v}_{\mathrm{b}}\parallel <0.02 \right]\land\mathbb{I} \left[ \parallel \boldsymbol{\omega }_{\mathrm{b}}\parallel <0.02 \right]\\
\end{array} \hspace{-5pt}\right]\\[10pt]
	\cdot \left[\hspace{-5pt}\begin{array}{c}
	e^{-\frac{\parallel \boldsymbol{p}_e\parallel}{0.02}}+e^{-\frac{\parallel \boldsymbol{v}_{\mathrm{b}}\parallel}{0.02}}\\
	+e^{-2\frac{\parallel \boldsymbol{\theta }_e\parallel}{0.05}}+e^{-2\frac{\parallel \boldsymbol{\omega }_{\mathrm{b}}\parallel}{0.02}}\\
\end{array}\hspace{-5pt}\right]\\
\end{array} $ & $0.75$\\
    \midrule[0.7pt]
    \rowcolor[gray]{0.9} \multicolumn{3}{l}{\textbf{\textit{Regularization}}} \\
    \midrule[0.7pt]
    Linear Velocity & $\left( e^{0.6\cdot \min \left( \left\| \boldsymbol{v}_{\mathrm{b}} \right\| ,4.0 \right)}-1 \right) ^2$ & $ -0.03$ \\ 
    Angular Velocity & $\left( e^{0.4\cdot \min \left( \left\| \boldsymbol{\omega }_{\mathrm{b}} \right\| ,6.0 \right)}-1 \right)^2$  & $-0.05$\\
    Joint Limitation & $\sum_i^4{\left( \left(\left| q_{i}^{*} \right|-\bar{q}\right)\mathbb{I}\left[\left| q_{i}^{*} \right|>\bar{q}\right) \right]}$ & $-0.02$
    \\
    Thrust Limitation & $\sum_i^4{\left( f_{i}^{*}-\bar{f} \right)^2 \mathbb{I}\left[ f_{i}^{*} > \bar{f} \right]}  $ & $-0.01$
    \\
    Joint Action Rate & $\sum_i^4{\left( q_{i,t}^{*}-q_{i,t-1}^{*} \right) ^2}$  & $-5.0\times 10^{-4}$
    \\
    Thrust Action Rate & $\sum_i^4{\left( f_{i,t}^{*}-f_{i,t-1}^{*} \right) ^2}$  & $-1.0\times 10^{-4}$
    \\
    Thrust Power & $\sum_i^4{f_{i}^{2}}$  & $-1.0\times 10^{-5}$\\
    Joint Acceleration & $\sum_i^4{\left( \dot{q}_{i,t}-\dot{q}_{i,t-1} \right)^2}$  & $-1.5\times 10^{-5}$
    \\
    z-Axis Align & $\mathbb{I}\left[|b_{\mathrm{z}}-b_{\mathrm{z}}^{*}|>1.5\right]$ & $-0.1$
    \\
    Thrust Allocation & $\sum_i^4{\left( f_{i,t}^{*}-\frac{\sum_j^4{f_{j,t}^{*}}}{4} \right)^2}b_{\mathrm{z}}^2$ & $-1.0\times 10^{-5}$
    \\
    Early Termination & ground collision or $p_{\mathrm{w},z} > 6.0$ & $-1.0$\\
	\hline
	\end{tabular}
    }
    \vspace{-0.1cm}
\end{table}

Here, $\mathbb{I}\left[\cdot\right]$ is the indicator function. $\boldsymbol{p}_e = \boldsymbol{p}_{\mathrm{b}} - \boldsymbol{p}_{\mathrm{b}}^{*}$ denotes the position error in body frame and $\boldsymbol{p}_{\mathrm{b}}$ is the current body position. $\boldsymbol{\theta}_e \in \mathbb{R}^3$ denotes the orientation error between the current and desired base frames in axis-angle form, computed as $\mathrm{Log}(\mathbf{q}_{\mathrm{b}}^{*} \otimes \mathbf{q}_{\mathrm{b}}^{-1})$. $b_{\mathrm{z}}, b_{\mathrm{z}}^{*} \in [-1, 1]$ are the projections of the body $z$-axes of the current and desired base frames onto the world $z$-axis, providing a compact measure of the body tilt error through their difference.

\textit{Towards Position} encourages the robot to move toward the target position, while \textit{Towards Pose} promotes orientation alignment as the position error decreases. \textit{Reach Position} and \textit{Reach Pose} provide sparse bonuses once the robot enters the specified thresholds and remains stable. \textit{Linear Velocity} and \textit{Angular Velocity} penalties discourage aggressive motion and promote stable flight. Unlike the pose-reaching formulations commonly adopted in legged robot learning~\cite{ben2025gallant}, we avoid explicit time- or reference-velocity-based terms that enforce convergence within a prescribed duration; instead, the reward design balances task completion, energy efficiency, and motion smoothness, allowing the policy to autonomously discover control strategies that achieve agile yet stable pose convergence. Additional regularization terms penalize actuator-limit violations, abrupt action changes, excessive thrust power, and large joint accelerations. The \textit{z-Axis Align} auxiliary reward promotes reaching the desired tilt angle. \textit{Thrust Allocation} encourages an even distribution of rotor thrusts across varying base orientations, thereby balancing actuator workloads. \textit{Early Termination} suppresses inefficient exploration of failure episodes by aborting upon ground collision or exceeding the altitude bound.

\subsubsection{Initial State Distribution}\label{sec:initialState}
The desired position $\boldsymbol{p}_{\mathrm{w}}^{*}\in\mathbb{R}^3$ is drawn uniformly within a \SI{4}{\metre} $\times$ \SI{4}{\metre} $\times$ \SI{3}{\metre} region and transformed into the body frame as $\boldsymbol{p}_{\mathrm{b}}^{*}$. The desired orientation $\boldsymbol{\theta}_{\mathrm{b}}^{*}$ is sampled to broadly cover $SO(3)$ by first drawing the target body $+z$-axis on the unit sphere and then applying a yaw rotation uniformly in $[-\pi, \pi]$ about it. The robot is initialized at the world origin with zero linear and angular velocities, default hovering thrust $\mathrm{C}_{\mathrm{f}}$, and orientation sampled in the same manner of $\boldsymbol{\theta}_{\mathrm{b}}^{*}$. The initial joint position $\boldsymbol{q}_{\mathrm{j},0}$ requires particular care, since static hovering is feasible only at specific joint--thrust configurations. Initializing with default (fixed) joint configurations tends to converge to local optima, while uniform random sampling can lead to inefficient convergence. To encourage exploration while maintaining convergence efficiency, $\boldsymbol{q}_{\mathrm{j},0}$ is stratified across parallel environments into four equally weighted groups: (i) all joints at $-\pi$, (ii) all at $0$, (iii) all at $\pi$, and (iv) each joint independently sampled from $\{-\pi, 0, \pi\}$.

\subsection{Sim-to-Real Transfer}\label{sec:sim2real}
\subsubsection{System Latency}
System latency is unavoidable on real-world platforms and can degrade control performance or even induce divergent oscillations. To bridge this gap, we explicitly model the dominant delays during training: the commanded $\boldsymbol{q}_{\mathrm{j},t}^{*}$ and $\boldsymbol{f}_{\mathrm{r},t}^{*}$ are applied to the simulated actuators with an actuation delay, and the corresponding joint feedback $\boldsymbol{q}_{\mathrm{j},t}$ is returned with a sensing delay. The base velocities $\boldsymbol{v}_{\mathrm{b},t}$ and $\boldsymbol{\omega}_{\mathrm{b},t}$ further carry an algorithmic delay from the EKF state estimator, which fuses motion-capture and IMU measurements. These delays are estimated by collecting real platform data.

\subsubsection{Rotor Model}\label{sec:rotorModule}
The rotor module serves as the primary actuator of the aerial robot. Since rotor thrust dynamics are highly nonlinear and depend on the coupled effects of motor characteristics, propeller aerodynamics, and supply voltage, accurate first-principles modeling is impractical. We instead identify the rotor thrust and reaction torque empirically by sweeping PWM commands and supply voltages on a 6-axis force-torque sensor, and fit polynomial functions to map the PWM command $\mathrm{cmd}_{\mathrm{r}}$ and voltage $V_{\mathrm{r}}$ to the generated thrust and torque:
\begin{equation}
\boldsymbol{f}_{\mathrm{r}} = \mathrm{f}_{\mathrm{r}}(\mathrm{cmd}_{\mathrm{r}}, V_{\mathrm{r}}), \quad \boldsymbol{\tau}_{\mathrm{r}} = C_{\Omega}\boldsymbol{f}_{\mathrm{r}},
\end{equation}
where $C_{\Omega}$ denotes the torque-to-thrust ratio. During training, the commanded thrust is realized as $\boldsymbol{f}_{\mathrm{r},t} = \boldsymbol{f}_{\mathrm{r},t}^{*}$ and applied as external force and reaction torque on each rotor link. During deployment, the PWM command is recovered through the inverse mapping $\mathrm{cmd}_{\mathrm{r},t} = \mathrm{f}_{\mathrm{r}}^{-1}(\boldsymbol{f}_{\mathrm{r},t}^{*}, V_{\mathrm{r}})$.

\subsubsection{Joint Model}\label{sec:jointModule}
Joint modeling is critical for tilt-rotor quadrotors, since the rotor-generated wrench is steered by the joint positions, distinguishing them from fixed-rotor platforms. Rotational joints in common rigid-body simulators are modeled as DC motors with proportional-derivative (PD) position control:
\begin{equation}
    J \ddot{q}_{\mathrm{j}} = K_p (q_{\mathrm{j}}^* - q_{\mathrm{j}}) - K_d \dot{q}_{\mathrm{j}},
\end{equation}
where $J$ is the effective rotational inertia, and $K_p$, $K_d$ are the stiffness and damping coefficients, respectively. The resulting closed-loop joint dynamics are well-approximated by a second-order transfer function
\begin{equation}
    \frac{q_{\mathrm{j}}(s)}{q_{\mathrm{j}}^*(s)} = \frac{K_p}{Js^2 + K_d s + K_p},
\end{equation}
with natural frequency $\omega_n$ and damping ratio $\zeta$ satisfying
\begin{equation}
    \omega_{n}^{2} = K_p/J, \quad 2\zeta\omega_n = K_d/J.
\end{equation}
We identify $\omega_n$ and $\zeta$ from the measured joint response, then back-compute $K_p$ and $K_d$ for the simulated joints via the relations above. The joint torque and velocity limits are similarly determined experimentally. During deployment, the joint position commands are directly issued as $\mathrm{cmd}_{\mathrm{q},t} = \boldsymbol{q}_{\mathrm{j},t}^{*}$.
\begin{table}[htbp]
\vspace{-0.5cm}
\setlength{\tabcolsep}{4pt}
\caption{Domain Randomization, Noise and Time Delay}\label{tab:randomization}
\vspace{-0.1in}
\begin{center}
\begin{tabular}{lcc}
\toprule[1.0pt]
\textbf{Variables} & \textbf{Operation} & \textbf{Latency} $(\mathrm{d}t=\SI{10}{\milli\second})$ \\
\midrule[0.5pt]

\rowcolor[gray]{0.9} \multicolumn{3}{l}{\textbf{\textit{Dynamics Parameters}}} \\
\midrule[0.5pt]

mass [\si{\kilogram}] 
& scale [0.9,1.1] & - \\
inertia [\si{\kilogram\metre\squared}] 
& scale [0.9,1.1] & - \\
COM [\si{\metre}] 
& bias $\pm 0.01$ & - \\
\midrule[0.5pt]
\rowcolor[gray]{0.9} \multicolumn{3}{l}{\textbf{\textit{State Observations}}} \\
\midrule[0.5pt]

$\boldsymbol{p}_{\mathrm{b}}$ [\si{\metre}] 
& bias $\pm 0.01$, noise $\pm 0.01$ & 0 \\
$\boldsymbol{\theta}_{\mathrm{b}}$ [\si{\deg}] 
& bias $\pm 0.01$, noise $\pm 0.01$ & 0 \\
$\boldsymbol{v}_{\mathrm{b}}$ [\si{\metre\per\second}]
& bias $\pm 0.005$, noise $\pm 0.005$ & 3$\sim$5 \\
$\boldsymbol{\omega}_{\mathrm{b}}$ [\si{\radian\per\second}] 
& bias $\pm 0.005$, noise $\pm 0.005$ & 3$\sim$5 \\
$\boldsymbol{q}_{\mathrm{j}}$ [\si{\deg}] 
& bias $\pm 3$, noise $\pm 2$ & 1$\sim$2 \\
\midrule[0.5pt]
\rowcolor[gray]{0.9} \multicolumn{3}{l}{\textbf{\textit{Action Commands}}} \\
\midrule[0.5pt]

$\boldsymbol{a}_{\mathrm{j}}$ [\si{\radian}]
& - & 1$\sim$3 \\
$\boldsymbol{a}_{\mathrm{r}}$ [\si{\newton}] 
& - & 1$\sim$3 \\

\bottomrule[1.0pt]
\end{tabular}
\end{center}
\vspace{-0.5cm}
\end{table}

\begin{table*}[!b]
\vspace{-0.3cm}
\caption{\textbf{Continuous Pose Reaching in Simulation}}\label{tab:1000sample}
\vspace{-0.1in}
\begin{center}
\setlength{\tabcolsep}{3pt} 
\resizebox{\textwidth}{!}{
\begin{tabular}{lcccccc}
\toprule[1.0pt]
\multicolumn{1}{l}{} & 
$\mathcal{U}\!\parallel \boldsymbol{p}_e\!\parallel$\cilmit{std,min,max}[\si{\metre}] $\downarrow$ & $\mathcal{U}\!\parallel \boldsymbol{\theta }_e\!\parallel$\cilmit{std,min,max}[\si{\deg}] $\downarrow$ & 
$\mathcal{L}\!\parallel \boldsymbol{p}_e\!\parallel$\cilmit{std,min,max}[\si{\metre}] $\downarrow$ & $\mathcal{L}\!\parallel \boldsymbol{\theta }_e\!\parallel$\cilmit{std,min,max}[\si{\deg}] $\downarrow$ & 
$\left| \boldsymbol{q}_{\mathrm{j}}^{*}-\boldsymbol{q}_{\mathrm{j}} \right|$\ci{std}[\si{\radian}] $\downarrow$ & $\boldsymbol{f}_{\mathrm{r}}^{*}$\ci{std}[\si{\newton}] $\downarrow$\\

\midrule[0.7pt]

RL      & 0.077\cilmit{0.029,0.019,0.304} & \textbf{3.88\cilmit{5.70,0.25,116.73}}
        & \textbf{0.127\cilmit{0.071,0.030,0.895}} & \textbf{7.85\cilmit{10.60,0.41,145.77}}
        & 0.029\ci{0.083} &10.00\ci{3.27}\\
\addlinespace[2pt]
NMPC    & \textbf{0.024}\cilmit{0.011,0.002,0.068} & 25.12\cilmit{35.06,0.16,170.78} 
        & 0.422\cilmit{0.147,0.096,0.955} & 40.93\cilmit{31.03,6.49,171.01}
        & \textbf{0.004}\ci{0.015} & \textbf{9.31}\ci{2.97}\\

\bottomrule[1.0pt]
\end{tabular}
}
\arrayrulecolor{black}
\end{center}
\end{table*}
\subsubsection{Domain Randomization and Noise}\label{sec:randomization}
Domain randomization (DR) is widely adopted to improve policy robustness and facilitate sim-to-real transfer, but excessive randomization trades off task-level performance~\cite{tan2018sim}. We therefore apply only minimal and physically grounded randomization to the body mass, body inertia, and joint position offsets, while injecting observation noise to account for sensor uncertainties. Combined with the actuator-level system identification of Sec.~\ref{sec:rotorModule} and Sec.~\ref{sec:jointModule}, this conservative scheme keeps the training environment closely aligned with the real platform while preserving robustness. The details of the proposed sim-to-real pipeline for the hybrid joint-rotor actuation system are summarized in Table~\ref{tab:randomization}.

\section{Experiments}\label{sec:experiments}

\subsection{Experimental Setup}\label{sec:setup}
The training environment is built on Isaac Sim~\cite{isaacsim2025} and Isaac Lab~\cite{isaaclab2025}, and the policy is optimized with proximal policy optimization (PPO)~\cite{schulman2017proximal} using the RSL-RL implementation~\cite{schwarke2025rslrl}. Training is performed on a single NVIDIA A10 GPU, consuming approximately \SI{16}{\hour} for 20,000 iterations; full hyperparameters are listed in Appendix~\ref{apd:training}. The learned policy is then deployed zero-shot in both simulation and on the real platform, without any fine-tuning. To facilitate reproducibility, the training and deployment code is publicly available at \href{https://github.com/ut-dragon-lab/aerial_lab}{github.com/ut-dragon-lab/aerial\_lab}.

As a baseline, we reuse the servo-integrated NMPC of~\cite{li2024servo}, running at \SI{100}{\hertz} on the onboard computer and fully integrated into the robot software stack as the control module analogous to the actor network in Fig.~\ref{fig:framework}. Its prediction model combines single-rigid-body dynamics with a first-order servo model over a horizon of \SI{2}{\second} (20 steps), and shares the thrust mapping of Sec.~\ref{sec:rotorModule}. The NMPC parameters are fine-tuned for our platform and held fixed across the different scenarios.
\subsection{Simulation Experiments}\label{sec:simulation}
We first evaluated the learned policy and the NMPC baseline in a standalone Gazebo simulation that mirrors the real-platform control architecture.

\subsubsection{Waypoint Hovering}\label{sec:simwaypoint}
\begin{figure}[htbp]
    \centering
    \vspace{-0.3cm}
    \includegraphics[width = 8.6 cm]{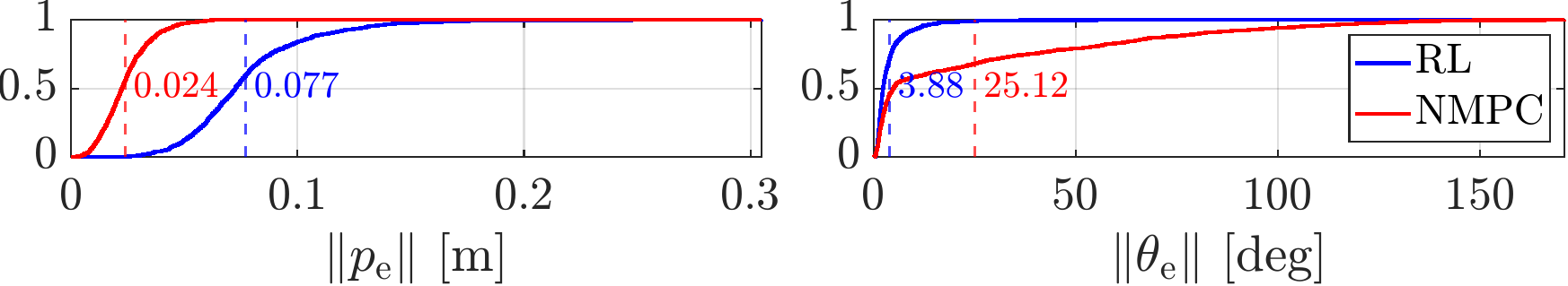}
    \caption{Cumulative distributions of steady-state position and orientation errors across the sampled target set.}\label{fig:sampleSS}
    \vspace{-0.4cm}
\end{figure}

We sampled 1,000 target poses in $SE(3)$ following the scheme of Sec.~\ref{sec:initialState} and commanded the robot to reach them sequentially, \SI{8}{\second} per pose. Results are summarized in Table~\ref{tab:1000sample} and Fig.~\ref{fig:sampleSS}, where $\mathcal{U}(\cdot)$ and $\mathcal{L}(\cdot)$ denote the average over the static-hovering window (\SI{0.5}{\second} before the next target, reflecting steady-state accuracy) and the transition window (\SI{6.0}{\second} at the middle of the horizon, reflecting transient agility), respectively. $\parallel \boldsymbol{p}_e\!\parallel$ and $\parallel \boldsymbol{\theta}_e\!\parallel$ represent the L2-norm of position and orientation error, respectively.

Both controllers could reach the sampled targets in $SE(3)$, with complementary strengths: NMPC achieves lower steady-state position error, while the RL policy yields a tighter orientation-error distribution (Fig.~\ref{fig:sampleSS}). The larger orientation error of NMPC arises from pose transitions: for certain target pairs the controller executes a body rotation that drives the joints into their position limits, leaving a residual orientation error that can no longer be corrected. The RL policy also converges faster during the transition phase, as reflected in the lower transient errors $\mathcal{L}\!\parallel\!\boldsymbol{p}_e\!\parallel$ and $\mathcal{L}\!\parallel\!\boldsymbol{\theta}_e\!\parallel$ in Table~\ref{tab:1000sample}.

\subsubsection{Ablation Studies}\label{sec:ablation}
We conducted ablation studies to evaluate the contribution of specific reward terms, as shown in Table~\ref{tab:ablation} (rows (a), (b), and (c) indicate the ablated settings).
(a) Removing \textit{Reach Position/Pose} leaves the steady-state pose errors essentially unchanged, but increases joint tracking error and action rates, indicating that this term mainly suppresses residual oscillations near the target.
(b) Removing \textit{z-Axis Align} elevates the transient orientation errors and the body velocity, confirming its role as an auxiliary orientation regularizer.
(c) Removing \textit{Thrust Allocation} increases the mean thrust and joint tracking error, undermining actuator reliability in real-world deployment.
Overall, these ablated terms act as soft regularizers rather than primary drivers of pose convergence: task-level rewards are sufficient for $SE(3)$ pose control, while these regularizers shape \textit{how} the dynamics are coordinated.
\begin{table}[htbp]
\vspace{-0.2cm}
\begin{center}
\caption{\textbf{Rewards Ablation Study Results}}\label{tab:ablation}
\setlength{\tabcolsep}{2.5pt} 
\resizebox{0.49\textwidth}{!}{
\begin{tabular}{lcccccccccc}
\toprule[1.0pt]
\addlinespace[-0.05pt]
\multicolumn{1}{l}{} & $\scriptstyle\mathcal{U}\!\parallel \boldsymbol{p}_e\!\parallel\,\downarrow$ & $\scriptstyle\mathcal{U}\!\parallel \boldsymbol{\theta }_e\!\parallel\,\downarrow$ &
$\scriptstyle\mathcal{L}\!\parallel \boldsymbol{p}_e\!\parallel\,\downarrow$ & $\scriptstyle\mathcal{L}\!\parallel \boldsymbol{\theta }_e\!\parallel\,\downarrow$ &
$\scriptstyle\mathcal{U}\!\parallel\boldsymbol{v}_{\mathrm{b}}\!\parallel\,\downarrow$ & $\scriptstyle\mathcal{U}\!\parallel \boldsymbol{\omega }_{\mathrm{b}}\!\parallel\,\downarrow $ &
$\scriptstyle\left| \boldsymbol{q}_{\mathrm{j}}^{*}-\boldsymbol{q}_{\mathrm{j}} \right|\,\downarrow$ & $\scriptstyle\boldsymbol{f}_{\mathrm{r}}^{*}\,\downarrow$ &
$\scriptstyle\left| \dot{\boldsymbol{q}}_{\mathrm{j},t}^{*} \right|\,\downarrow$ & $\scriptstyle\left|\dot{\boldsymbol{f}}_{\mathrm{r},t}^{*}\right|\,\downarrow$\\
\multicolumn{1}{l}{} & {\scriptsize [\si{\metre}]} & {\scriptsize [\si{\deg}]} & {\scriptsize [\si{\metre}]} & {\scriptsize [\si{\deg}]} & {\scriptsize [\si{\metre\per\second}]} & {\scriptsize [\si{\radian\per\second}]} & {\scriptsize [\si{\radian}]} & {\scriptsize [\si{\newton}]} & {\scriptsize [\si{\radian}/step]} & {\scriptsize [\si{\newton}/step]}\\
\noalign{\vskip -2pt}
\midrule[0.7pt]
Base    & 0.077 & 3.88
        & 0.127 & 7.85
        & 0.009 & 0.007 
        & 0.029 & 10.00
        & 0.69 & 7.35\\
(a)     & \textbf{\textcolor{gray}{0.081}} & \textbf{\textcolor{black}{3.79}}
        & \textbf{\textcolor{gray}{0.135}} & \textbf{\textcolor{black}{7.26}}
        & \textbf{\textcolor{gray}{0.010}} &  \textbf{\textcolor{gray}{0.010}}
        & \textbf{\textcolor{gray}{0.032}} & \textbf{\textcolor{gray}{10.22}}
        & \textbf{\textcolor{gray}{0.79}} & \textbf{\textcolor{gray}{7.77}}\\
(b)     & \textbf{\textcolor{gray}{0.081}} & \textbf{\textcolor{black}{3.83}}
        & \textbf{\textcolor{gray}{0.134}} & \textbf{\textcolor{gray}{7.96}}
        & \textbf{\textcolor{gray}{0.011}} & \textbf{\textcolor{gray}{0.009}}
        & 0.029 & \textbf{\textcolor{black}{9.93}}
        & 0.69  & \textbf{\textcolor{gray}{7.62}}\\
(c)     & \textbf{\textcolor{gray}{0.079}} & \textbf{\textcolor{black}{3.74}}
        & \textbf{\textcolor{gray}{0.140}} & \textbf{\textcolor{gray}{8.27}}
        & \textbf{\textcolor{gray}{0.012}} & \textbf{\textcolor{gray}{0.012}}
        & \textbf{\textcolor{gray}{0.031}} & \textbf{\textcolor{gray}{10.29}}
        & 0.69 & \textbf{\textcolor{black}{7.32}}\\
\addlinespace[-0.5pt]
\bottomrule[1.0pt]
\end{tabular}  
}
\arrayrulecolor{black}
\vspace{-0.8cm}
\end{center}
\end{table}

\subsection{Real-world Experiments}\label{sec:realworld}
The real platform carries a KHADAS VIM4 onboard computer (\SI{2.2}{\giga\hertz} quad-core Cortex-A73 + \SI{2.0}{\giga\hertz} quad-core Cortex-A53) running Ubuntu 20.04 with ROS Noetic for state estimation, policy inference, and NMPC optimization, while an STM32 board handles low-level control. Notably, the RL policy inference takes about \textbf{\SI{0.3}{\milli\second}}, while NMPC typically fits within the \SI{10}{\milli\second} control period but occasionally overruns. This headroom is beneficial for onboard deployment on resource-constrained aerial platforms. The control rate bounds the bandwidth of the disturbances and actuator dynamics that a controller can compensate. Nonetheless, we ran both controllers at \SI{100}{\hertz}, within the current \SI{100}{\hertz} joint-command update rate (Fig.~\ref{fig:framework}) and to ensure a fair comparison.

\subsubsection{Waypoint Hovering}\label{sec:realwaypoint}
The robot was commanded to sequentially reach a set of target poses with a period of \SI{8}{\second}, and the test was repeated three times. Quantitative metrics are summarized in Table~\ref{tab:waypoints}, with snapshots shown in Fig.~\ref{fig:highlight}(a). For reference, RMSE results of similar tests from~\cite{li2024servo} are X: \SI{0.072}{\metre}, Y: \SI{0.029}{\metre}, Z: \SI{0.044}{\metre}, (norm \SI{0.089}{\metre}) Roll: \SI{3.250}{\deg}, Pitch: \SI{2.810}{\deg}, Yaw: \SI{4.342}{\deg}, (norm \SI{6.108}{\deg}) which are comparable to the results in Table~\ref{tab:waypoints}.

\begin{figure}[htbp]
    \vspace{-0.3cm}
    \centering
    \subfigure[Position Trajectory]{
        \centering
        \includegraphics[width=8.4cm]{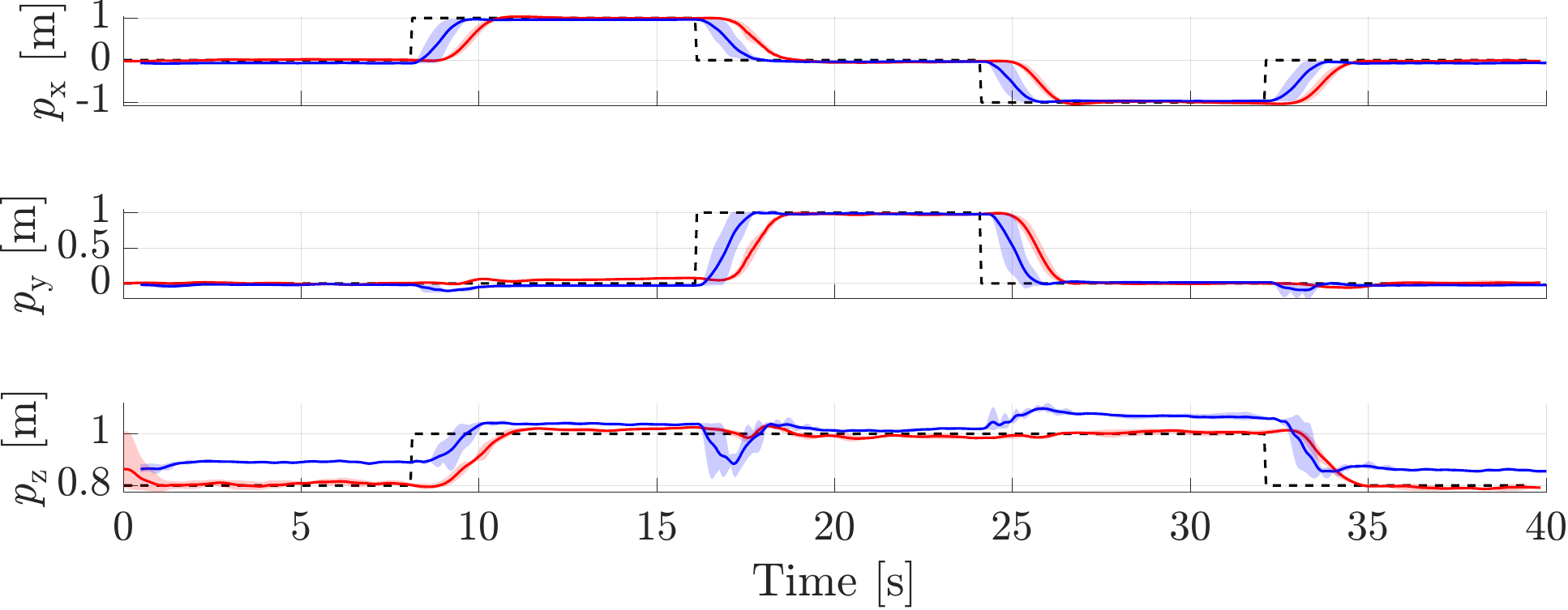}
        \label{fig:waypointsPos}
    }\\ 
    \vspace{-0.2cm}
    \subfigure[Orientation Trajectory]{
        \centering
        \includegraphics[width=8.4cm]{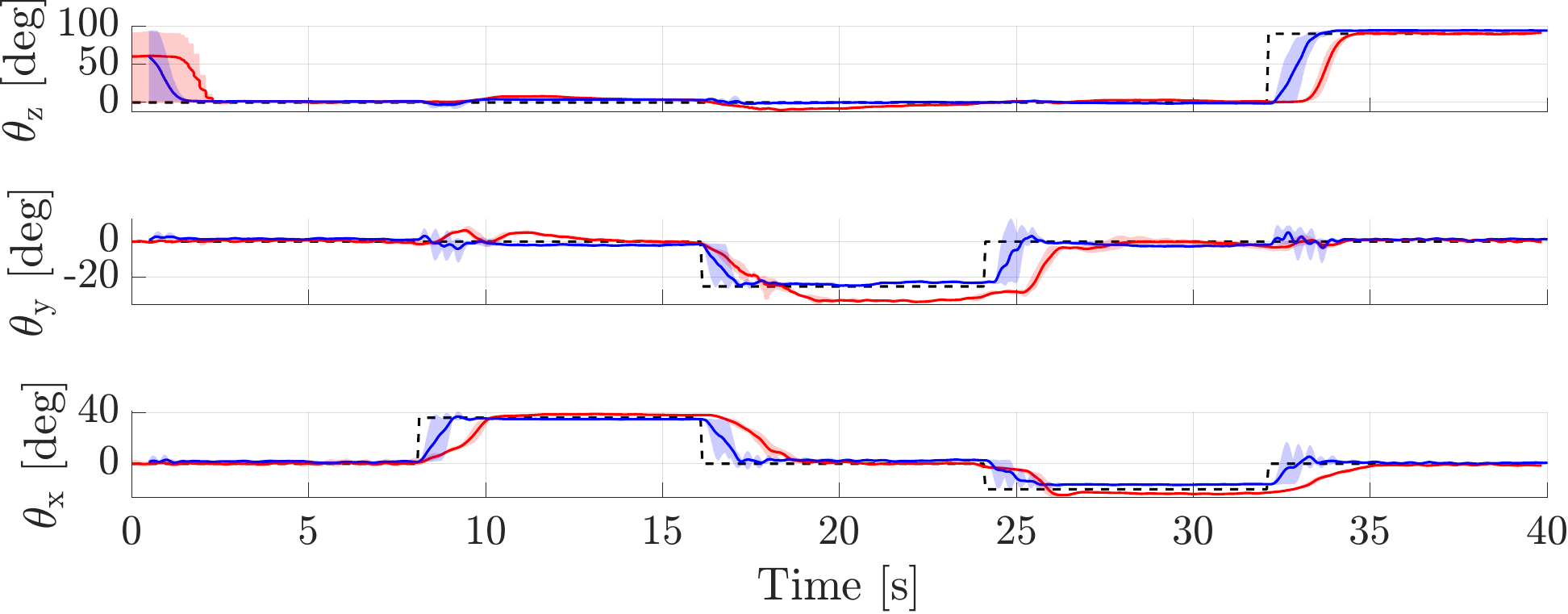}
        \label{fig:waypointsAng}
    }
    \vspace{-0.2cm}
    \caption{State trajectories from real-world waypoint hovering experiments. Solid lines show the mean values across trials, and the shaded regions represent the corresponding minimum and maximum ranges.}
    \label{fig:waypointtraj}
    \vspace{-0.4cm}
\end{figure}

\begin{table}[htbp]
\vspace{-0.2cm}
\setlength{\tabcolsep}{3pt}
\caption{\textbf{Continuous Pose Reaching in Real World}}\label{tab:waypoints}
\vspace{-0.1in}
\begin{center}
\resizebox{0.48\textwidth}{!}{
\begin{tabular}{cccccc}
\toprule[1.0pt]
\multicolumn{1}{c}{\multirow{2}{*}{$\boldsymbol{p}_{\mathrm{w}}^{*}$[\si{\metre}]}} & \multicolumn{1}{c}{\multirow{2}{*}{$\boldsymbol{\theta}_{\mathrm{b}}^{*}$[\si{\deg}]}} & \multicolumn{2}{c}{\textit{RL}} & \multicolumn{2}{c}{\textit{NMPC}} \\
\cmidrule[0.7pt](lr){3-4} \cmidrule[0.7pt](lr){5-6} 

\multicolumn{2}{c}{} & 
$\mathcal{U}\!\parallel \boldsymbol{p}_e\!\parallel$[\si{\metre}] $\downarrow$ & $\mathcal{U}\!\parallel \boldsymbol{\theta }_e\!\parallel$[\si{\deg}] $\downarrow$ & 
$\mathcal{U}\!\parallel \boldsymbol{p}_e\!\parallel$[\si{\metre}] $\downarrow$ & $\mathcal{U}\!\parallel \boldsymbol{\theta }_e\!\parallel$[\si{\deg}] $\downarrow$ \\
\midrule[0.7pt]

[0.0, 0.0, 0.8] & [0, 0, 0] 
& 0.114 & \textbf{2.589}
& \textbf{0.026} & 2.644 \\[0.4ex]

[1.0, 0.0, 1.0] & [45, 0, 0]
&  \textbf{0.060} & \textbf{4.350} 
& 0.074 & 4.372 \\[0.4ex]

[0.0, 1.0, 1.0] & [0, -25, 0]
& \textbf{0.046} & \textbf{4.214} 
& 0.054 & 8.225 \\

[-1.0, 0.0, 1.0] & [-25, 0, 0]
& 0.074 & 5.561 
& \textbf{0.028} & \textbf{4.742} \\  [0.4ex]

[0.0, 0.0, 0.8] & [0, 0, 90]
& 0.091 & 4.590 
& \textbf{0.028} & \textbf{1.768} \\  [0.4ex]

\midrule[0.7pt]

\multicolumn{2}{c}{\cellcolor[gray]{0.9}Mean}
& \cellcolor[gray]{0.9}0.077 & \cellcolor[gray]{0.9} \textbf{4.261} 
& \cellcolor[gray]{0.9}\textbf{0.042} & \cellcolor[gray]{0.9}4.350 \\

\bottomrule[1.0pt]
\end{tabular}
}
\vspace{-0.5cm}
\end{center}
\end{table}
The RL policy achieves position and orientation accuracy comparable to NMPC across the five real-world targets with more uniform per-target orientation errors, consistent with observations in Sec.~\ref{sec:simwaypoint}. Additionally, the state trajectories shown in Fig.~\ref{fig:waypointtraj} further show faster pose-error convergence under RL, with peak velocities exceeding \SI{2}{\metre\per\second} and \SI{5}{\radian\per\second}. This transient behavior is consistent with recent comparative studies of learned and classical quadrotor controllers~\cite{kunapuli2025leveling}

\subsubsection{Disturbance Rejection}\label{sec:disturbance}
\begin{figure}[htbp]
    \centering
    \vspace{-0.3cm}
    \includegraphics[width=8.4cm]{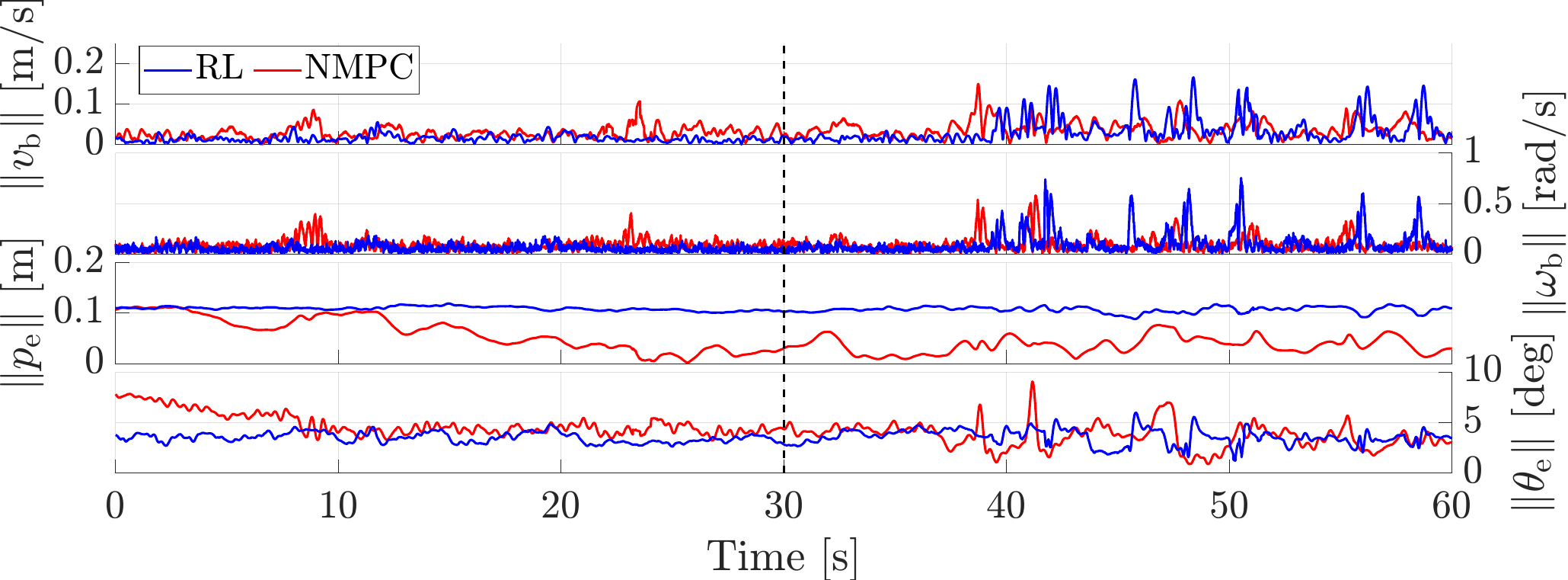}
    \caption{Real-world disturbance rejection evaluation. Left: response under wind. Right: response to impulsive stick pushes.}
    \label{fig:disturbance}
    \vspace{-0.3cm}
\end{figure}

We evaluated disturbance rejection at the target pose $\boldsymbol{p}_{\mathrm{w}}^{*}=[0.0,\,0.0,\,1.0]$~\si{\metre}, $\boldsymbol{\theta}_{\mathrm{b}}^{*}=[0,\,0,\,0]$~\si{\deg} under two disturbance conditions: a continuous \SI{2.5}{\metre\per\second} wind generated by a fan (Fig.~\ref{fig:highlight}(f)) and impulsive pushes applied to the robot body (Fig.~\ref{fig:highlight}(e)). The resulting pose errors and body velocities are reported in Fig.~\ref{fig:disturbance}.

Under wind, the RL policy yields smaller position and orientation fluctuations than NMPC, with reduced standard deviations of \SI{0.005}{\metre} vs.\ \SI{0.028}{\metre} and \SI{0.63}{\deg} vs.\ \SI{1.32}{\deg}. It does show a larger mean steady-state position offset (\SI{0.10}{\metre} vs.\ \SI{0.04}{\metre}) but a smaller mean orientation error (\SI{3.55}{\deg} vs.\ \SI{4.27}{\deg}), consistent with the trade-off observed in waypoint hovering. Under impulsive pushes, the RL policy recovers more rapidly, exhibiting sharper but shorter velocity peaks and lower average linear and angular speeds (\SI{0.025}{\metre\per\second} vs.\ \SI{0.033}{\metre\per\second}; \SI{0.081}{\radian\per\second} vs.\ \SI{0.092}{\radian\per\second}). Under both wind and impulsive disturbances, the RL policy exhibits substantially smaller fluctuations and faster recovery than NMPC, with a trade-off of a larger steady-state position offset.

\begin{figure}[htbp]
    \vspace{-0.2cm}
    \centering
    \includegraphics[width = 8.4 cm]{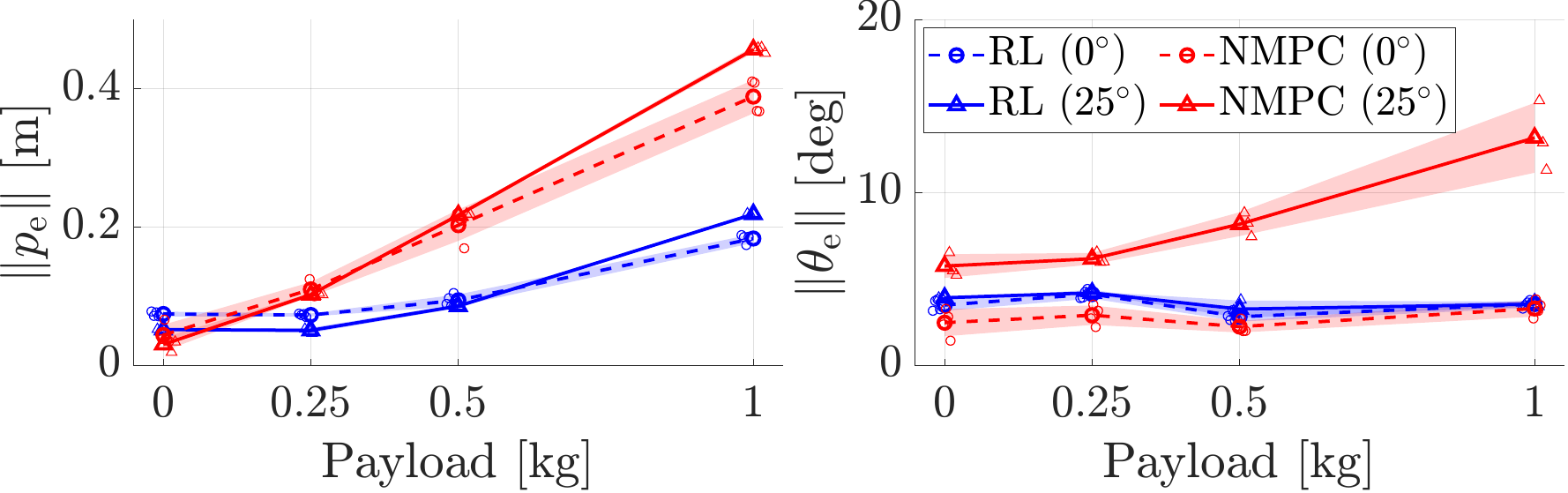}
    \caption{Steady-state position and orientation errors under varying payloads. Lines denote the mean over all trials, shaded regions the min--max range, and markers the individual samples. Each evaluation trial contains four samples of nominal pose and three samples of tilted pose.}
    \label{fig:weightsexp}
    \vspace{-0.6cm}
\end{figure}

We further evaluated payload variation by attaching additional weights, as shown in Fig.~\ref{fig:highlight}(d). The robot alternated between $\boldsymbol{p}_{\mathrm{w}}^{*}\!=\![0.0,\,0.0,\,1.0]$~\si{\metre}, $\boldsymbol{\theta}_{\mathrm{b}}^{*}\!=\![0,\,0,\,0]$~\si{\deg} and $\boldsymbol{p}_{\mathrm{w}}^{*}\!=\![0.25,\,0.0,\,0.8]$~\si{\metre}, $\boldsymbol{\theta}_{\mathrm{b}}^{*}\!=\![0,\,25,\,0]$~\si{\deg}, repeated three nominal--tilted pose cycles for each of the four payload conditions. As expected, the added mass mainly inflates the position error through the higher effective gravitational load; NMPC starts from a lower unloaded error but degrades about twice as fast, reaching about \SI{0.4}{\metre} vs.\ \SI{0.2}{\metre} at \SI{1}{\kilogram} (Fig.~\ref{fig:weightsexp}). Notably, the RL orientation error stays below \SI{5}{\deg} at both target poses, whereas NMPC rises to about \SI{13}{\deg} at the \SI{25}{\deg} pitch pose. A \SI{1}{\kilogram} payload is a $33\%$ mass change, well beyond the $\pm10\%$ mass randomization used in training. Yet the policy generalizes to this out-of-distribution load without fine-tuning, preserving its orientation regulation.
\subsection{Extended Studies}\label{sec:extended}
\subsubsection{Omnidirectional Operation}\label{sec:omnidirectional}
\begin{figure}[htbp]
    \centering
    \vspace{-0.3cm}
    \includegraphics[width = 8.6 cm]{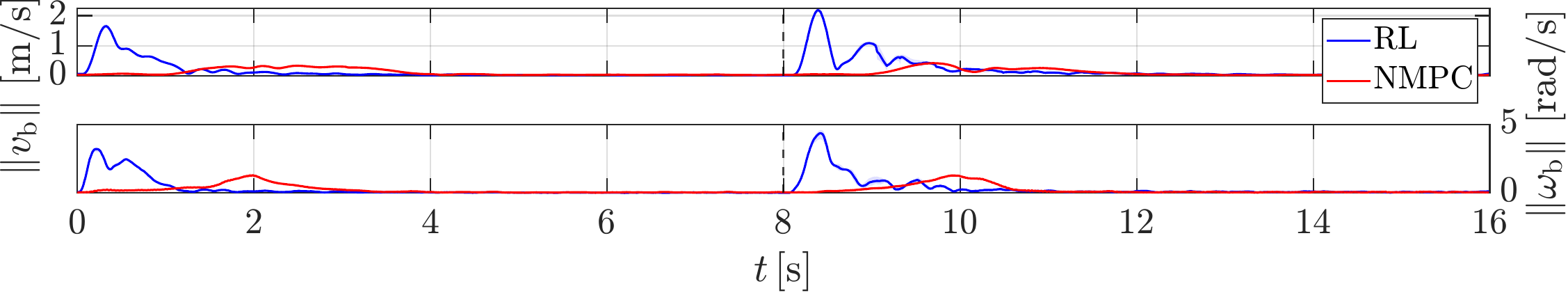}
    \caption{Velocity profile of the \SI{90}{\deg} pitch maneuver.}\label{fig:90pitch}
    \vspace{-0.4cm}
\end{figure}

We verified $SE(3)$ pose control under large-tilt configurations, which require highly coupled thrust maneuvering, through a real-world hovering test (Fig.~\ref{fig:highlight}(c), Fig.~\ref{fig:90pitch}). The platform was reconfigured for this test (battery removed and protective structure added for safety concerns, total mass \SI{2.56}{\kilogram}), and the policy was retrained under the same framework with only the inertial parameters updated, demonstrating the generality of the proposed framework. The robot flew from $\boldsymbol{p}_{\mathrm{w}}^{*}\!=\![0.0,\,0.0,\,0.8]$~\si{\metre}, $\boldsymbol{\theta}_{\mathrm{b}}^{*}\!=\![0,\,0,\,0]$~\si{\deg} to $\boldsymbol{p}_{\mathrm{w}}^{*}\!=\![0.5,\,0.0,\,1.0]$~\si{\metre}, $\boldsymbol{\theta}_{\mathrm{b}}^{*}\!=\![0,\,90,\,0]$~\si{\deg}, repeated three times. The policy achieves $\mathcal{U}\!\parallel\!\boldsymbol{p}_e\!\parallel=\!0.044$~\si{\metre} and $\mathcal{U}\!\parallel\!\boldsymbol{\theta}_e\!\parallel=\!5.18$~\si{\deg}, while NMPC demonstrates $\mathcal{U}\!\parallel\!\boldsymbol{p}_e\!\parallel=\!0.066$~\si{\metre} and $\mathcal{U}\!\parallel\!\boldsymbol{\theta}_e\!\parallel=\!9.09$~\si{\deg} with updated physical parameters. The velocity profiles (Fig.~\ref{fig:90pitch}) show that the policy reaches its peak velocities in shorter bursts, consistent with the faster pose transitions observed in the waypoint hovering tests, confirming the policy's uniform performance across diverse target poses in $SE(3)$.


\subsubsection{Trajectory Tracking}
\begin{figure}[htbp]
    \vspace{-0.3cm}
    \begin{minipage}{8.4cm} 
        \centering
        \includegraphics[width=8.4cm]{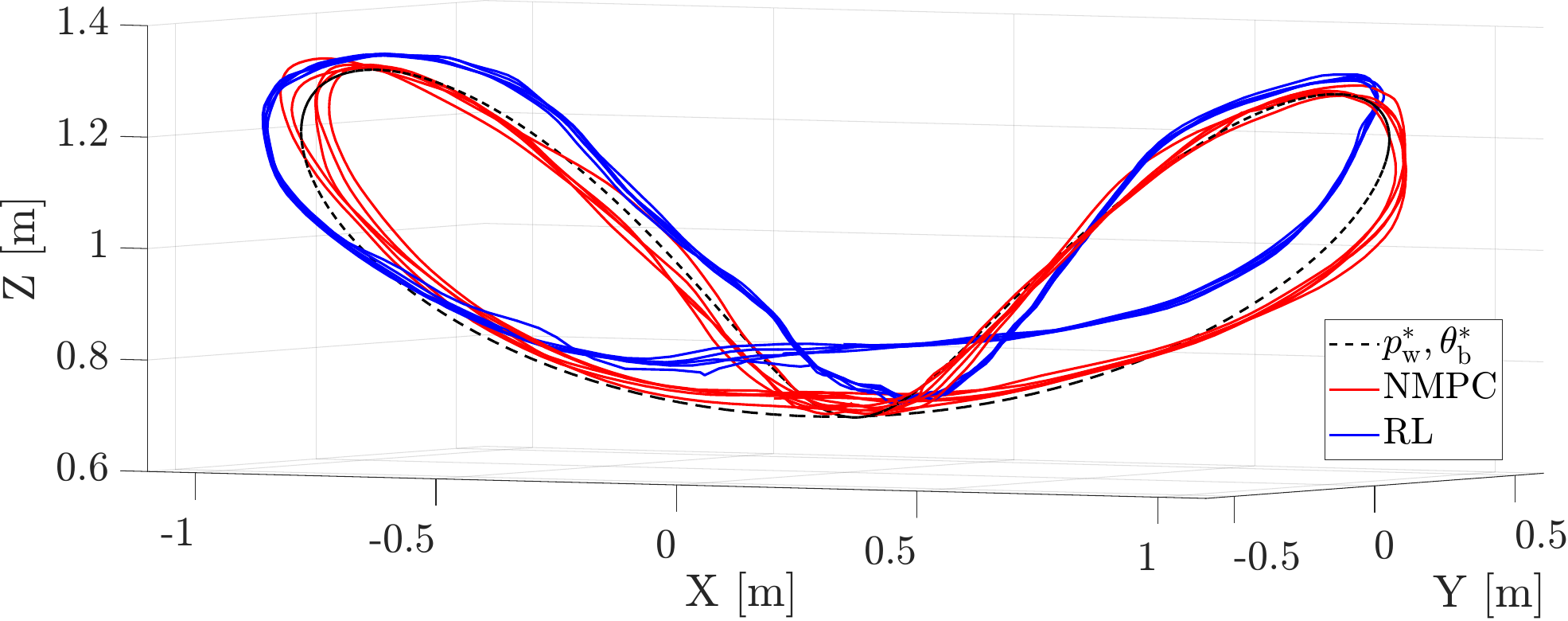}\\
        \vspace{0.1cm} 
        \includegraphics[width=8.4cm]{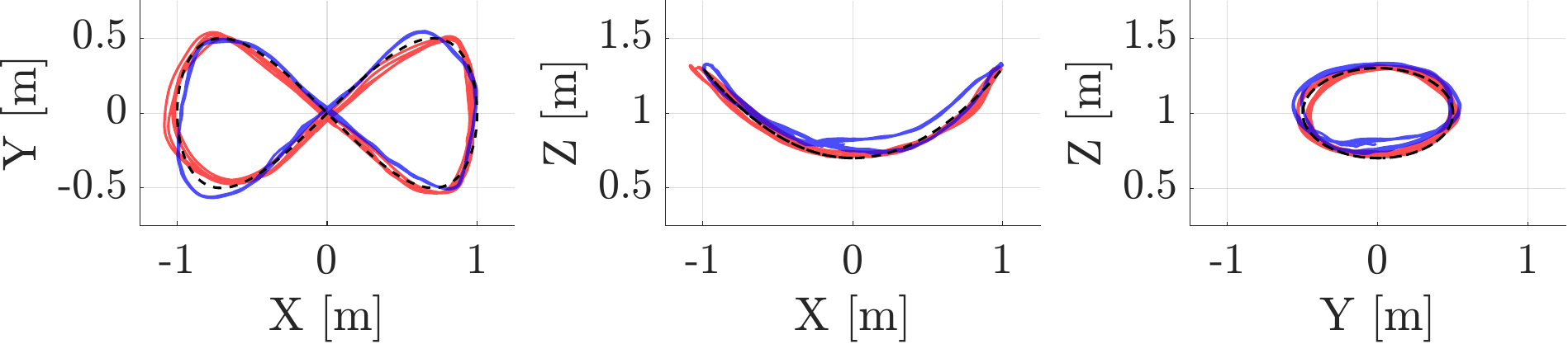}
    \end{minipage}
    \caption{Real-world trajectory tracking of a 3D lemniscate with synchronized orientation modulation, repeated five times.}
    \label{fig:tracking}
    \vspace{-0.3cm}
\end{figure}

Although the policy was trained only for pose-reaching, we further evaluated its generalization to trajectory tracking in the real world (Fig.~\ref{fig:highlight}(b)). The reference is a time-parameterized 3D lemniscate with synchronized orientation modulation (roll and pitch within \SI{0.5}{\radian}, yaw in $[0, 2\pi]$~\si{\radian}). The task was repeated five times, with results shown in Fig.~\ref{fig:tracking}. For reference, NMPC's RMSE errors for similar scenarios reported in~\cite{li2024servo} are X: \SI{0.071}{\metre}, Y: \SI{0.067}{\metre}, Z: \SI{0.018}{\metre}, (norm \SI{0.099}{\metre}); Roll: \SI{4.934}{\deg}, Pitch: \SI{2.023}{\deg}, Yaw: \SI{2.789}{\deg}, (norm \SI{6.018}{\deg}).

The RL policy successfully follows the trajectory despite never being trained for it. Its mean errors are larger than NMPC (\SI{0.16}{\metre} vs.\ \SI{0.05}{\metre} in position; \SI{8.06}{\deg} vs.\ \SI{5.19}{\deg} in orientation), but the tracking variability is comparable, with similar standard deviations in position (\num{0.04} vs.\ \num{0.02}) and orientation (\num{2.45} vs.\ \num{2.46}). This indicates that the learned policy generalizes to time-varying references beyond its training distribution, although high-precision tracking remains a strength of model-based control.

\subsubsection{Singularity Avoidance}\label{sec:singular}
\begin{figure}[htbp]
    \centering
    \vspace{-0.2cm}
    \includegraphics[width=\linewidth]{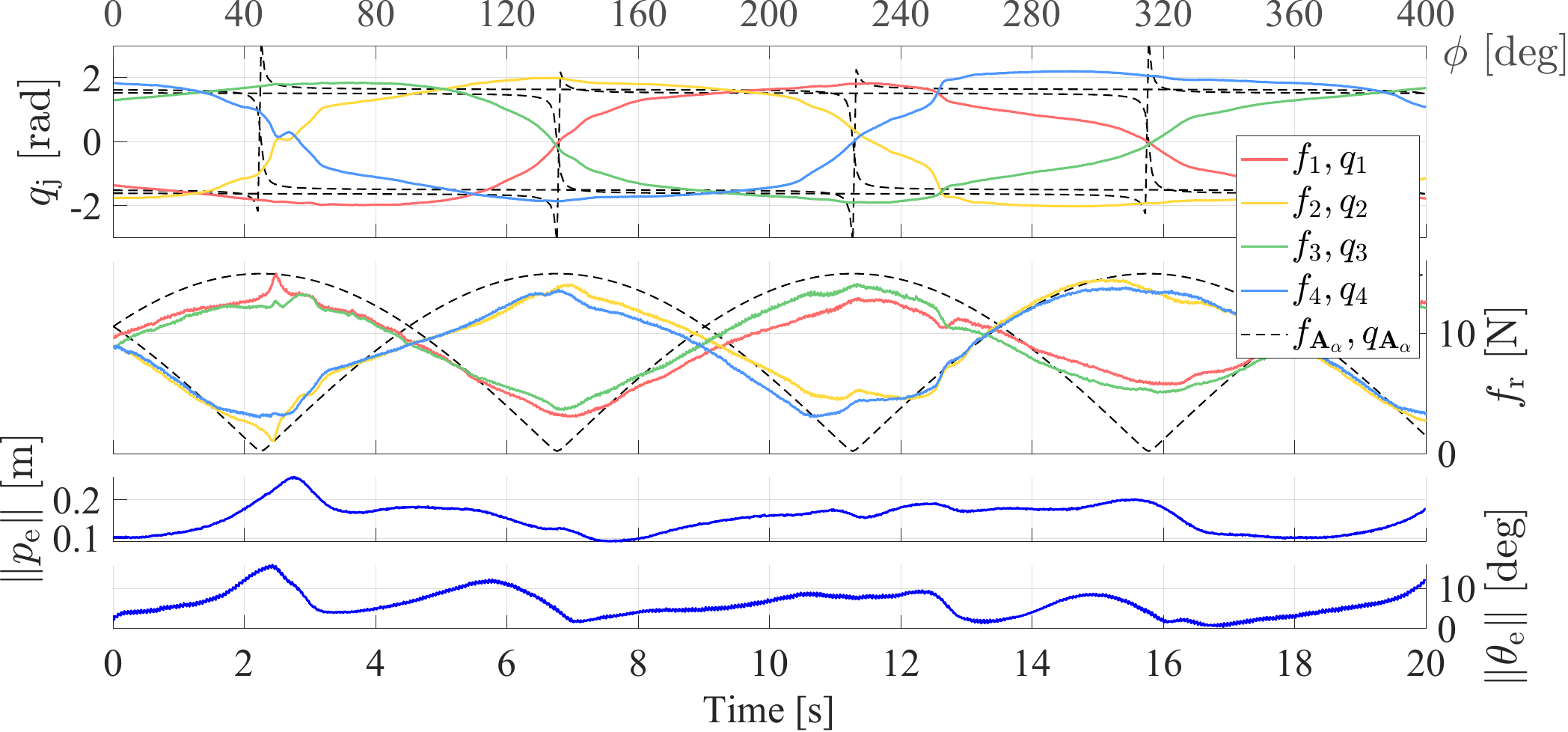}
    \caption{State evolution under continuous rotation through $\boldsymbol{A}_{\alpha}$ singularities.}\label{fig:rotation}
    \vspace{-0.5cm}
\end{figure}

We evaluated continuous rotation tracking in simulation, with target $\boldsymbol{p}_{\mathrm{w}}^{*}\!=\![0.0,\,0.0,\,1.2]$~\si{\metre} and $\boldsymbol{\theta}_{\mathrm{b}}^{*}\!=\![0,\,90,\,\phi]$~\si{\deg} (XYZ Euler angles), where $\phi$ sweeps continuously through $\boldsymbol{A}_{\alpha}$ singularities in vertical configurations (ill-conditioned near \SI{45}{\deg}, \SI{135}{\deg}, \ldots). As shown in Fig.~\ref{fig:rotation}, the learned policy produces continuous joint trajectories $\boldsymbol{q}_{\mathrm{j}}$ with pose errors remaining low. This also illustrates that, as a byproduct of actuator-level end-to-end learning, the policy naturally handles configurations that would otherwise require explicit allocation regularization~\cite{cuniato2026allocation}.

\section{Conclusion}\label{sec:conclusion}
This paper presented a sim-to-real reinforcement learning pipeline for $SE(3)$ pose-reaching on overactuated tilt-rotor quadrotors. A single policy directly produces joint and rotor commands for zero-shot real-world deployment with actuator-level system identification, latency alignment, and physically grounded domain randomization to bridge the gap induced by the hybrid joint-rotor actuation. Experiments demonstrate consistent performance across waypoint hovering, external disturbances, payload variation, trajectory tracking, and continuous rotation through allocation-singular configurations, the last typically requiring explicit regularization in allocation-based controllers. As the dynamics calculation is performed during training, the policy keeps its online computation minimal, which suits platforms with limited onboard resources and tasks that demand consistent behavior across conditions rather than absolute accuracy. Two limitations remain: the platform model omits aerodynamic effects such as wind drag and inter-rotor downwash, calling for richer real-to-sim modeling; and the reward relies on manually shaped terms to coordinate the joint-rotor dynamics, motivating more principled formulations. Further hardware optimization and control problem formulation are needed to fully unlock the potential of actuator-level policies on tilt-rotor platforms.

\vspace{-0.2cm}
\section*{Appendix} 
\subsection{Physical Parameters}\label{apd:physical}
\begin{table}[H]
\vspace{-0.3cm}
\setlength{\tabcolsep}{4pt}
\caption{Physical Parameters}\label{tab:physical}
\vspace{-0.1in}
\begin{center}
\resizebox{0.42\textwidth}{!}{
\begin{tabular}{lclc}
\toprule[1.0pt]
\textbf{Term} & \textbf{Value} & \textbf{Term} & \textbf{Value} \\
\midrule[0.5pt]

\rowcolor[gray]{0.9} \multicolumn{4}{l}{\textbf{\textit{Weight Parameters}}} \\
\midrule[0.5pt]
    mass [\si{\kilogram}] & 3.0386 & inertia $I_{x},I_{y},I_{z}$ [\si{\gram\metre\squared}] & 63, 62, 95 \\
\midrule[0.5pt]
\rowcolor[gray]{0.9} \multicolumn{4}{l}{\textbf{\textit{Joint Parameters}}} \\
\midrule[0.5pt]
    $K_p$ &  0.3449 &  $K_d$ & 0.0094\\
    position limit [\si{\radian}]& 3.96 & velocity limit [\si{\radian\per\second}]& 10.0\\
    effort limit [\si{\newton\metre}]& 3.0 & & \\
\midrule[0.5pt]
\rowcolor[gray]{0.9} \multicolumn{4}{l}{\textbf{\textit{Rotor Parameters}}} \\
\midrule[0.5pt]
    thrust limit $\bar{f}$ [\si{\newton}] & 20 & hovering thrusts $\mathrm{C}_{\mathrm{f}}$ [\si{\newton}] & 7.0 \\
    torque-to-thrust ratio $C_{\Omega}$ & 0.05 &  &  \\
\bottomrule[1.0pt]
\end{tabular}
}
\end{center}
\vspace{-0.5cm}
\end{table}
\subsection{Training Hyperparameters}\label{apd:training}
\begin{table}[htbp]
\vspace{-0.3cm}
\setlength{\tabcolsep}{4pt}
\caption{Hyperparameters}\label{tab:hyperparameters}
\vspace{-0.1in}
\begin{center}
\resizebox{0.42\textwidth}{!}{
\begin{tabular}{lclc}
\toprule[1.0pt]
\textbf{Term} & \textbf{Value} & \textbf{Term} & \textbf{Value} \\
\midrule[0.5pt]

\rowcolor[gray]{0.9} \multicolumn{4}{l}{\textbf{\textit{General}}} \\
\midrule[0.5pt]
    num of robots       & 4096  & num of epochs & 5 \\
    gradient clipping   & 1.0   & adam epsilon & $1.0e-8$ \\
    num of steps per iteration & 48 & num of steps per episode & 1,200 \\
    \midrule[0.5pt]
    \multicolumn{2}{l}{total training iterations} & \multicolumn{2}{c}{20,000} \\
\midrule[0.5pt]
\rowcolor[gray]{0.9} \multicolumn{4}{l}{\textbf{\textit{PPO}}} \\
\midrule[0.5pt]
    clip range &  0.2 &  entropy coefficient  & $5.0e-4$\\
    discount factor $\gamma$ & 0.99 & GAE balancing factor $\lambda$ & 0.95 \\
    desired KL-divergence & 0.01 & activation function & \textit{elu} \\
    \midrule[0.5pt]
    \multicolumn{2}{l}{actor neural network} & \multicolumn{2}{c}{MLP, hidden units [512, 256, 128]} \\
    \multicolumn{2}{l}{critic neural network} & \multicolumn{2}{c}{MLP, hidden units [512, 256, 128]} \\
    \multicolumn{2}{l}{output layer} & \multicolumn{2}{c}{Unbounded linear} \\
\midrule[0.5pt]
\rowcolor[gray]{0.9} \multicolumn{4}{l}{\textbf{\textit{Specific}}} \\
\midrule[0.5pt]
    joint action scale $c_{\mathrm{j}}$ & 0.25 & rotor action scale $c_{\mathrm{r}}$ & 0.5 \\
\bottomrule[1.0pt]
\vspace{-0.7cm}
\end{tabular}
}
\end{center}
\vspace{-0.3cm}
\end{table}



\ifCLASSOPTIONcaptionsoff
  \newpage
\fi



\bibliographystyle{IEEEtran}
\bibliography{IEEEabrv,references}

\end{document}